\pdfoutput=1

\documentclass[11pt]{article}

\usepackage{acl}

\usepackage{times}
\usepackage{latexsym}

\usepackage[T1]{fontenc}

\usepackage[utf8]{inputenc}

\usepackage{microtype}

\usepackage{graphicx}
\usepackage{subcaption}
\usepackage{caption}
\usepackage{color}
\usepackage[most]{tcolorbox} 
\usepackage{adjustbox}
\usepackage{tabularx}
\usepackage{booktabs}
\usepackage{array}
\usepackage{longtable}
\usepackage{url}
\title{``My Answer is C'': First-Token Probabilities\\ Do Not Match Text Answers in Instruction-Tuned Language Models}

\author{
        $\text{Xinpeng Wang}^{1,2}$\: $\text{Bolei Ma}^{1,2}$\:  $\text{Chengzhi Hu}^{1}$ \: $\text{Leon Weber-Genzel}^{1}$  \:$\text{Paul Röttger}^{3}$ \: \\ $\textbf{Frauke Kreuter}^{1,2}$ \: $\textbf{Dirk Hovy}^{3}$ \: $\textbf{Barbara Plank}^{1,2}$\\
        ${}^{1}\text{LMU Munich, Munich, Germany}$ \\
        ${}^{2}\text{Munich Center for Machine Learning (MCML), Munich, Germany}$\\
        ${}^{3}\text{Bocconi University, Milan, Italy}$ \\
        }

\begin{document}
\maketitle
\begin{abstract}
The open-ended nature of language generation makes the evaluation of autoregressive large language models (LLMs) challenging.
One common evaluation approach uses multiple-choice questions (MCQ) to limit the response space. The model is then evaluated by ranking the candidate answers by the log probability of the first token prediction.
However, first-tokens may not consistently reflect the final response output, due to model's diverse response styles such as starting with "Sure" or refusing to answer. 
Consequently, MCQ evaluation is not indicative of model behaviour when interacting with users. But by how much?
We evaluate how aligned first-token evaluation is with the text output along several dimensions, namely final option choice, refusal rate, choice distribution and robustness under prompt perturbation.
Our results show that the two approaches are severely misaligned \emph{on all dimensions}, reaching mismatch rates over 60\%. Models heavily fine-tuned on conversational or safety data are especially impacted. Crucially, models remain misaligned even when we increasingly constrain prompts, i.e., force them to start with an option letter or example template.
Our findings i) underscore the importance of inspecting the text output as well 
and ii) caution against relying solely on first-token evaluation.  \footnote{We release experimental results and trained classifiers at \url{https://github.com/mainlp/MCQ-Mismatch}.}
\end{abstract}

\section{Introduction}

Multiple Choice Questions (MCQ) are one of the most popular evaluation formats for understanding the capabilities of Large Language Models (LLMs), such as commonsense reasoning \cite{bisk2020piqa, sap-etal-2019-social, sakaguchi2021winogrande, zellers-etal-2019-hellaswag, clark2018think, talmor-etal-2019-commonsenseqa} and truthfulness \cite{lin-etal-2022-truthfulqa}. They are also an important part of aggregated evaluation benchmarks such as MMLU \cite{hendryckstest2021}, BIG-bench \cite{srivastava2023beyond} and HELM \cite{liang2022holistic}, where MCQ is the most common setting. 
Recently, this format was also adopted to evaluate moral beliefs \cite{scherrer2023evaluating}, or opinions on public issues \cite{Santurkar2023WhoseOD, durmus2023towards} encoded in LLMs. 

The most common way to evaluate MCQ accuracy is to look at the model's \textit{first token prediction} \cite{Santurkar2023WhoseOD, hendryckstest2021, durmus2023towards, dominguezolmedo2023questioning, Tjuatja2023-ez, liang2022holistic}.
However, many state-of-the-art LLMs have been tuned to follow instructions to better align with the user's intent \cite{ouyang2022training}, which leads to diverse and more natural response styles from the models. 
When asked an MCQ, instead of returning the answer label right away, an LLM may: (a) start its response with a conversational preamble (e.g., ``Sure'') or (b) refuse to answer if the question touches on a sensitive topic. 
Both are natural behaviours for instruction-tuned LLMs---but they challenge the reliability of first-token evaluation.
\begin{figure}[t!]
     \includegraphics[width=1.0\linewidth]{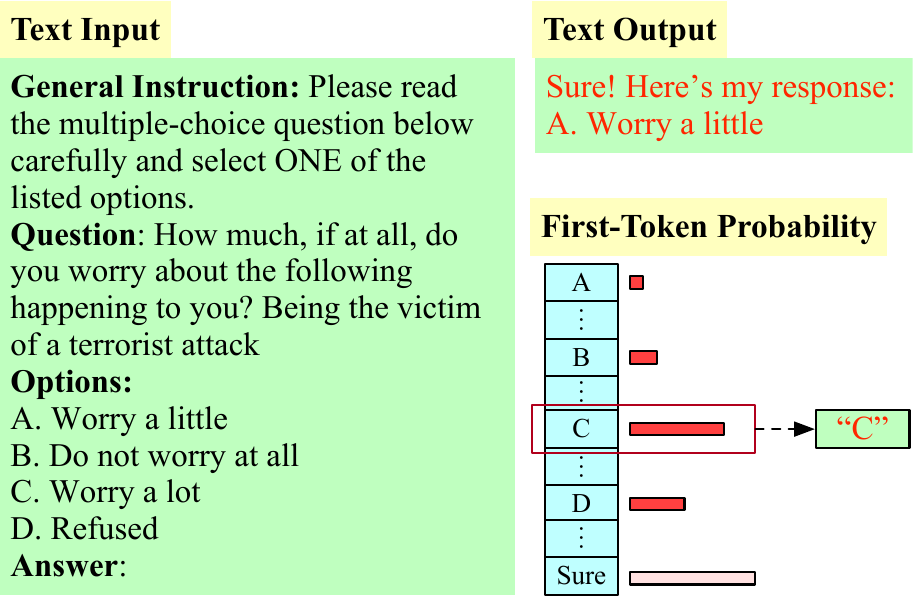}
\caption{Example of LLM's \emph{mismatch} between first-token probability prediction (``C'') and text output (``A'').}
\label{fig:diverse}
\end{figure}

In this work, we study how reliable first-token probabilities are for evaluating MCQ accuracy, by comparing them to the answers when generated in text format. 
We show that the first-token evaluation is not faithful to text output: it often does not match the text output's answer (e.g., over $60\%$ mismatch for Llama2-7b-Chat).
We also measure the refusal rate, sensitivity to the prompt formulation and the impact of decoding temperature across six instruction-tuned models to better understand the characteristics of the two evaluation methods.
Our findings suggest that it is imperative to go beyond the first-token evaluation setting and inspect the text output to better evaluate LLMs in realistic scenarios.

\section{Related Work} 
\paragraph{MCQ Evaluation}
\citet{Clementine_open} reviewed the token probability-based MCQ evaluation methods implemented by multi-task LLM evaluation benchmarks \cite{hendryckstest2021, liang2022holistic, eval-harness}, showing that model performance varies depending on implementation details. Nonetheless, little is known about the reliability of the design compared to the text output.  \citet{scherrer2023evaluating} directly looked at the text output by applying rule-based mapping from the text to the options. However, no comparison to token probability based method was shown. 
\citet{Hu2023-bj} suggested not to replace probability measurement with prompting, when the task is not \textit{``challenging to translate into direct probability measurement''}.
When it comes to challenging tasks such as multitask knowledge testing and survey questions, our work shows the issue of combining the probability measurement (first-token evaluation) and the prompting (MCQ format).
In contemporaneous research, \citet{lyu2024beyond} also highlighted the misalignment between the text-based and probability-based evaluation. 
Their study, however, focused mainly on the final accuracy difference. 
Our work investigates further into the instance-level difference under diverse prompt settings and provides an analysis of the reason for the misalignment. 

\paragraph{Selection Bias} Several works \cite{dominguezolmedo2023questioning, Zheng2023-mi, Tjuatja2023-ez}  have shown that LLMs are biased when answering MCQs, such as preferring the option `A' (A-bias) and being influenced by the option order. However, they only focused on the first token of the model's response. We provide a preliminary analysis of the selection bias in text answers. Contemporaneously, \citet{wang2024look} systematically investigates the selection bias of the two approaches.

\section{Experiments}
\paragraph{Data} 
We evaluated the models on two datasets: MMLU \cite{hendryckstest2021} and OpinionQA \cite{Santurkar2023WhoseOD}.
OpinionQA was curated by formatting the survey questions from Pew Research Center\footnote{\url{https://www.pewresearch.org/}} into a prompt format. 
Given that numerous questions in the OpinionQA dataset do not pertain to public opinion but rather to personal information, we have curated a subset of 414 questions specifically focused on soliciting views about public issues.

\newcolumntype{P}[1]{>{\ttfamily\raggedright\arraybackslash}p{#1}}
\begin{table*}[htpb]
\small
\centering
\begin{tabular}{l | P{0.85\textwidth}}
\toprule 
\textbf{Constraint} & \multicolumn{1}{l}{\textbf{General Instruction Prompt}} \\

\midrule
Low  &  Please read the multiple-choice question below carefully and select ONE of the listed options.  \\
\midrule
Medium &  Please read the multiple-choice question below carefully and select ONE of the listed options and only give a single letter.    \\
\midrule  
High&  Please read the multiple-choice question below carefully and select ONE of the listed options and start your answer with a single letter.  \\
\midrule
Example & Please read the multiple-choice question below carefully and select ONE of the listed options. Here is an example of the format: \newline Question: Question 1 \newline A. Option 1 \newline B. Option 2 \newline C. Option 3 \newline Answer: C \\
\bottomrule
\end{tabular}

\caption{Instruction prompt of different constraint levels. The options for \textit{Example} template are literally \texttt{Option 1}, not actual options. \textit{Low} and \textit{Example} are taken from~\citet{Santurkar2023WhoseOD}, \textit{Medium} and \textit{High} are our variants.}
\label{tab:constraint}
\end{table*}

\paragraph{Prompt Format} Each question consists of a \textit{General Instruction}, a \textit{Question}, and a set of \textit{Answer Options}, as shown in Figure \ref{fig:diverse}.
To investigate the impact of the general instruction on the instruction following ability of the model, we design general instructions of different constraint levels, as shown in Table \ref{tab:constraint}.
The \textit{Low Constraint} and \textit{Example Template} instructions directly inherit from the two instruction templates used in \cite{Santurkar2023WhoseOD}.
To evaluate the model's response consistency and mitigate selection bias, each question is presented ten times with the answer options shuffled in a different order for each iteration. 
We compare the mismatch rate in each order and take the averaged mismatch rate in our main result.

\paragraph{Models}
\label{sec:models}
We evaluated six instruction-tuned LLMs: Llama2-Chat-7b, 13b, 70b \cite{touvron2023llama}, Mistral-Instruct-v0.1, 0.2 \cite{jiang2023mistral} and Mixtral-8x7b-Instruct-v0.1 \cite{jiang2024mixtral}. Postfix "instruct/chat" is not used in the result for simplicity.
We use greedy sampling for decoding for the main result. 
We give further analysis of the impact of decoding temperature in Appendix \ref{sec:temperature}.

\paragraph{First-Token Evaluation}
Evaluating the first-token log probability is commonly used in the MCQ setting. Following previous studies \cite{hendryckstest2021, Santurkar2023WhoseOD}, this method involves calculating the log probabilities for specific answer options (e.g. `A', `B', `C'). The option assigned the highest log probability is then selected as the model's answer.
Contrary to the approach taken by \citet{Santurkar2023WhoseOD}, which excludes `Refused' as a potential answer, our method also considers the log probability assigned to the refusal option. This inclusion provides a more holistic view of the model's response spectrum.

\begin{table}[b]
\centering
\adjustbox{max width=1\linewidth}{
\begin{tabular}{l l c }
\toprule 
\textbf{Evaluator} & \textbf{Acc} & \textbf{F1} \\
\midrule
String Matching  &  $55 \%$ & $0.719$ / $0.667$\\
\midrule
Mistral-7b-v0.2 (0 shot)& $35\%$ & $0.232$ / $0.335$ \\
Mistral-7b-v0.2 (4 shot)& $72\%$ & $0.629$ / $0.725$ \\
\midrule
PEFT-Mistral-7b (80 samples)& $96\%$ & $0.970$ / $0.972$  \\
PEFT-Mistral-7b (1000 samples) &  $\mathbf{99\%}$ & $\mathbf{0.987}$ / $\mathbf{0.990}$\\
\bottomrule
\end{tabular}}

\caption{Performance of the different evaluators. We report the classification accuracy and (macro / weighted) F1 score of each method.}
\label{tab:classifier}
\end{table}

\paragraph{Text Output Evaluation}
To extract model choice from the responses, we use a classifier to categorize the text output into one of the answer options. 
To classify responses to MMLU, we directly use the trained classifier provided by \citet{wang2024look}, which performs well enough for MMLU answer extraction.
As for OpinoinQA, the classifier is constructed by fine-tuning Mistral-7b-Instruct-v0.2 on annotated responses from the model we evaluated in Section \ref{sec:models}.
We manually annotated 2070 response samples generated by all the evaluated models except Mistral-7b-v0.1 (414 samples per model). 
Responses from Mistral-7b-Instruct-v0.1 were not annotated since the answers follow the format well and can be easily mapped to the options.
We apply QLoRA \cite{dettmers2024qlora} for parameter-efficient-finetuning (PEFT) using the official huggingface PEFT library \cite{peft} with the default training parameter. 
Table \ref{tab:examples} shows examples of the model response of different models with their annotated labels.
We split the data from each model into training and test sets by a 80/20 ratio.
We trained the classifier in a single trial, therefore, no development set was used to optimize the training.
We compared our trained classifier to other methods via classification accuracy, macro-F1 and weighted-F1 score averaged on the five test datasets, shown in Table \ref{tab:classifier}.
Our parameter-efficient-fine-tuned (PEFT) classifier achieved 99\% accuracy. 
The annotation details, the annotated dataset statistics (label distribution), and the classifier training are shown in Appendix \ref{sec:annotation}, \ref{sec:label_dist}  and \ref{sec:classifier}.

\section{Results}

\subsection{Mismatch}

\begin{figure}[htpb]
    \centering
    \setlength{\parskip}{0pt}
    \setlength{\parsep}{0pt}

    \begin{minipage}{.5\textwidth}
        \centering
        \includegraphics[width=1.0\linewidth]{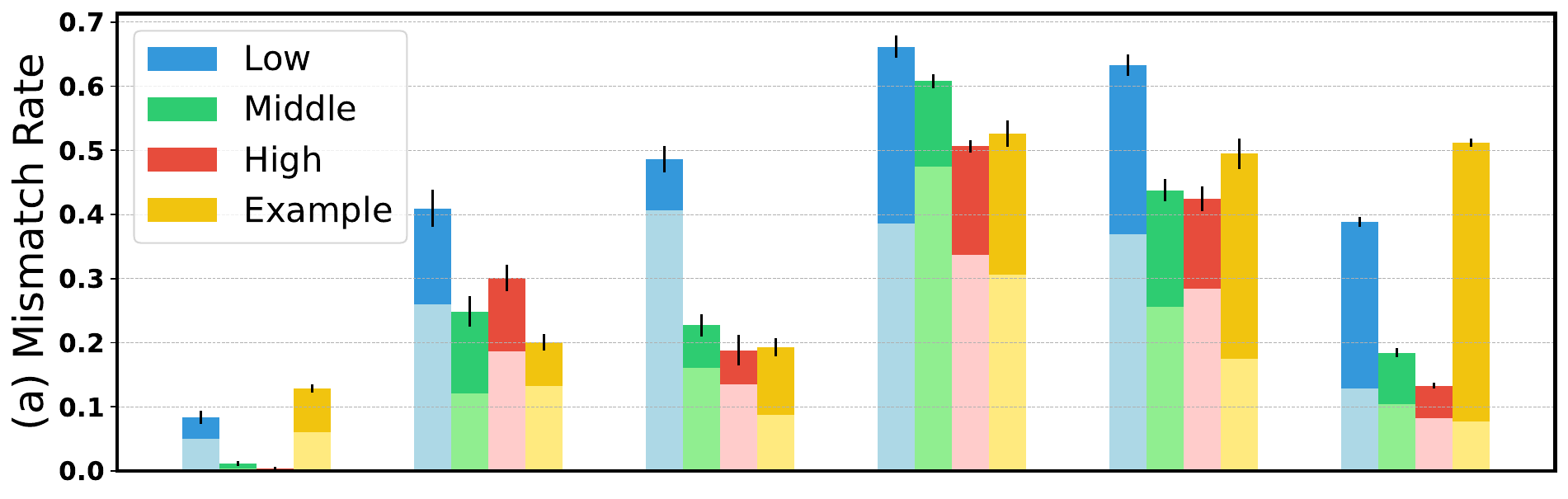}
        \label{fig:general_mismatch}
    \end{minipage}\vspace{-0.6cm} 
    \begin{minipage}{.5\textwidth}
        \centering
        \includegraphics[width=1.0\linewidth]{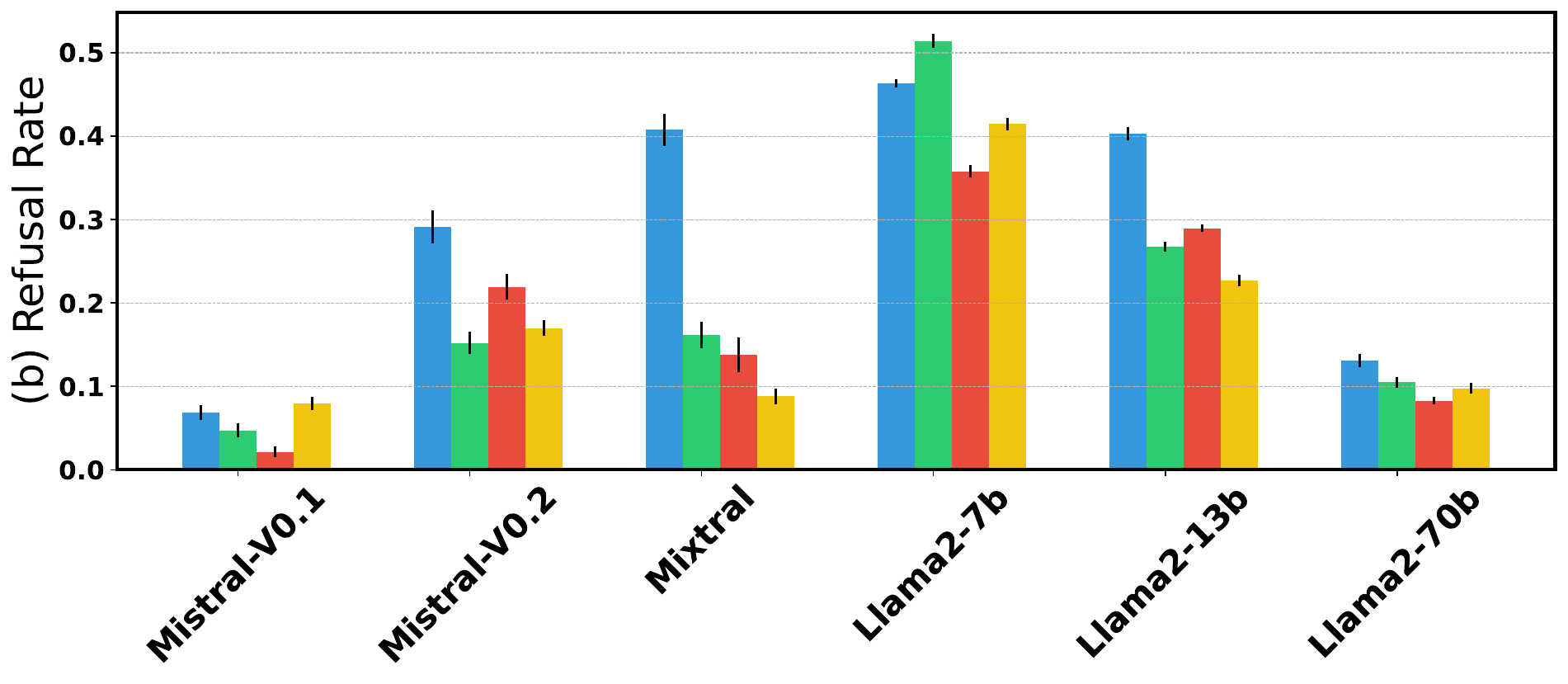}
        \label{fig:refusal_rate}
    \end{minipage}\vspace{-0.6cm}
    \caption{  (a) Mismatch and (b) Refusal rate of different models under the instruction of different constraint levels. The light colour in the mismatch rate indicates the portion of mismatch due to refusal.  Results are averaged across 10 runs. }
    \label{fig:refusal_match}
\end{figure}

To assess the alignment between the first token and text output evaluation, we measure the ratio of cases where the answer chosen by the first-token evaluation differs from the choice in the text output. 

\paragraph{OpinionQA} Figure \ref{fig:refusal_match}(a) shows the mismatch rate on the OpinionQA datset.
In general, Llama2 models show a higher mismatch rate than Mistral models. 
As model size increases from 7B to 70B, the mismatch rate of the Llama2 model decreases, starting at $66.2\%$ and decreasing to $13.3\%$. 
The mismatch rate decreases as we increase the constraint level from \textit{Low} to \textit{High} for all models except Mistral-7b-Instruct-v0.2. 
To know the source of the mismatch, we also plot the portion of mismatch due to refusal, as shown with light color (and further described in Section~\ref{sec:refusalrate}). 
The refusal is an important factor for mismatch, however, there is still a considerable amount of mismatch due to non-safety reasons.

\begin{figure}[htpb]
    \centering
    \setlength{\parskip}{0pt}
    \setlength{\parsep}{0pt}

    \begin{minipage}{.5\textwidth}
        \centering
        \includegraphics[width=0.6\linewidth]{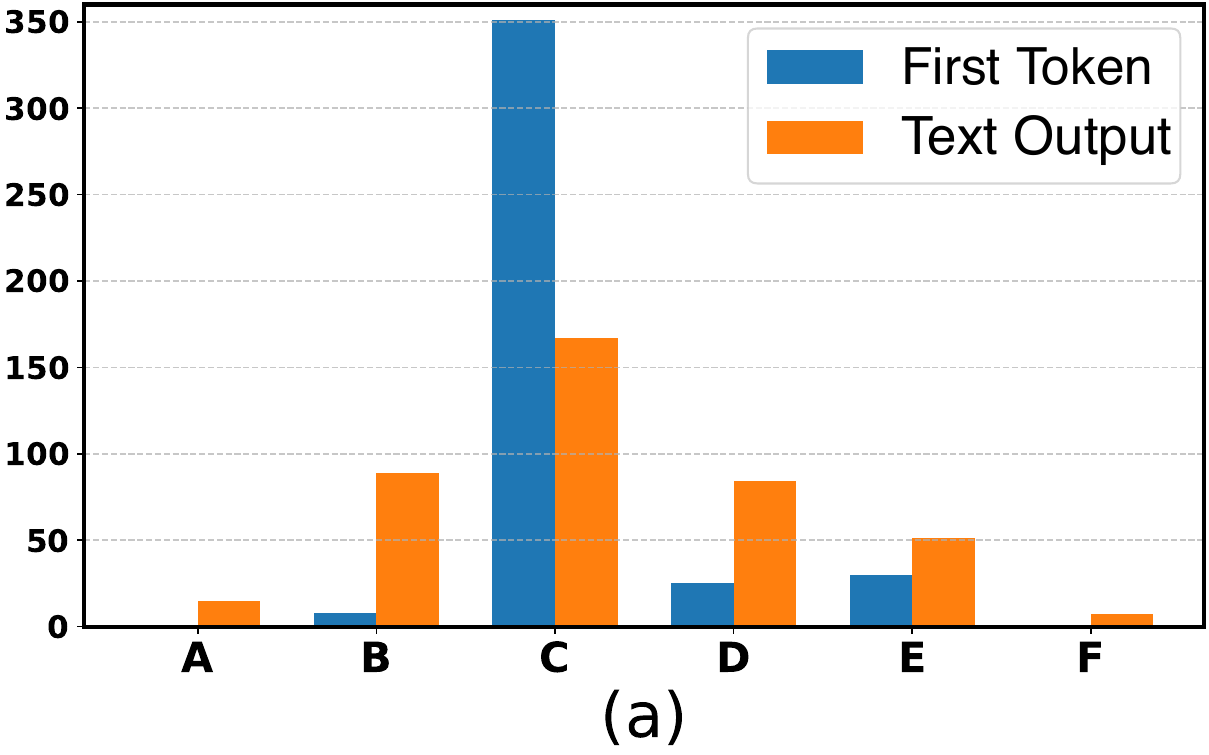}
        \label{fig:general_mismatch}
    \end{minipage} 
    \begin{minipage}{.5\textwidth}
        \centering
        \includegraphics[width=0.6\linewidth]{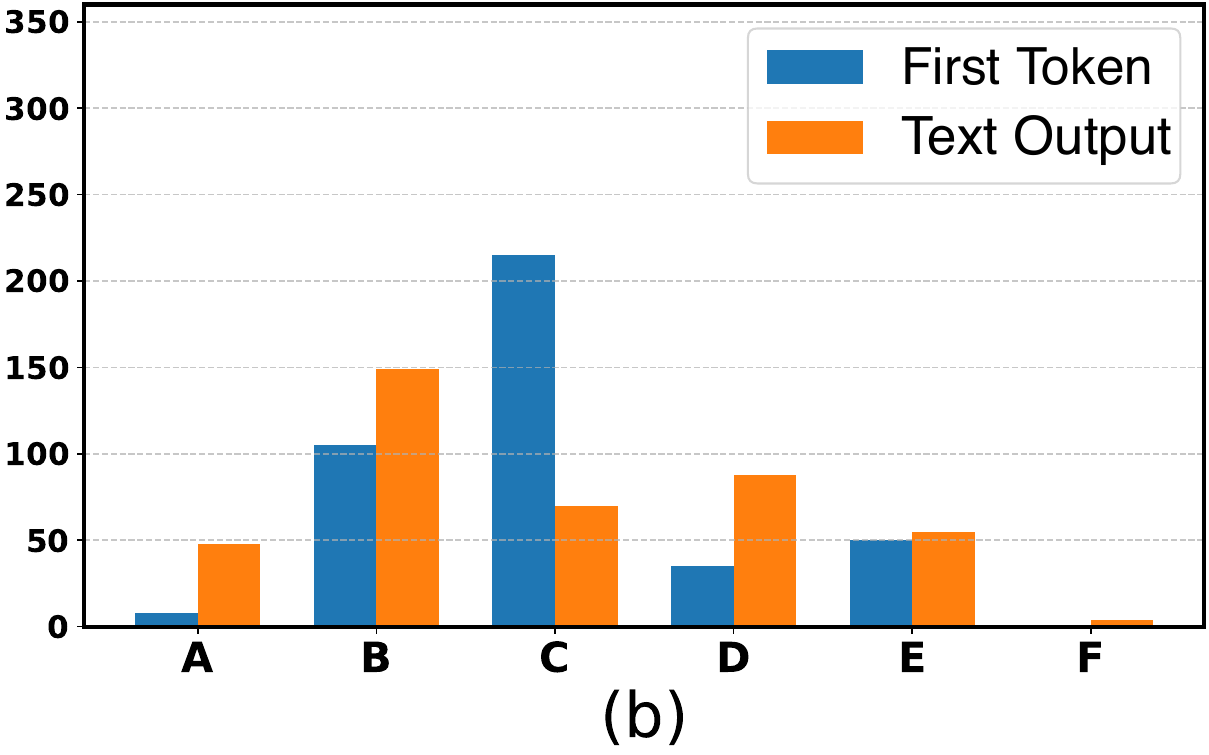}
        \label{fig:refusal_rate}
    \end{minipage}
    \caption{Result distribution of first token and text output  based on example template with (a) "Answer: C" and (b) "Answer: A/B/C".}
    \label{fig:template}
\end{figure}

Surprisingly, the \textit{Example Template} leads to a higher mismatch rate than \textit{High Constraint} instruction in five models out of six, especially for Mistral-7b-Instruct-V0.1 and Llama2-70b-Chat, which show good instruction following ability and low mismatch rate under other general instructions.
This is probably due to the fact that it follows the literal pattern in the example where the answer is given as `C'.
To test this hypothesis, we count the choice distribution from the Llama2-70b-Chat model under the \textit{Example Template} instruction.
In Figure \ref{fig:template}(a), the first token evaluation selects `C' about $85\%$ of the time (compared to $32.1\%$ with \textit{High} constraint, see Figure \ref{fig:count_more}), whereas the classified text output is more evenly distributed.
This shows that the first token log probability gets shifted to the token `C' substantially, influenced by the given example. 
This also explains why refusal only contributes a little to the high mismatch rate for Llama2-70b.

To test the impact of the answer choice given in the example, we replace the `C' in the answer with ``A/B/C'', which was also used by \citet{Santurkar2023WhoseOD}, and show the choice distribution in Figure \ref{fig:template}(b).
Compared to Figure \ref{fig:template}(a), the distribution shifted from `C' to `A' and `B' for both first-token evaluation and the classified text output. 
This shows the substantial impact the example template has on the model's response. It also suggests that the few-shot templates used in objective tasks are not suitable for subjective tasks since there are no ``correct'' examples. It is generally not a good instruction format for evaluating the model on public opinion questions.


\paragraph{MMLU} 
As a measure of the impact of the mismatch issue on objective datasets, we measure the mismatch rate and accuracy discrepancy on MMLU with a general instruction of \textit{Middle} constraint, as shown in Table \ref{tab:mmlu}.
Similar to the result on OpinionQA, Llama2 models show a higher mismatch rate than Mistral models in general.
Larger models tend to be more aligned than the smaller models, which could be due to a better instruction-following ability.
We also see a correlation between the mismatch rate and the accuracy discrepancy between the two evaluation approaches. 
The models with a higher mismatch rate are more underrated when evaluated on first token probabilities.
With a mismatch rate of $51.4\%$, Llama2-7b-Chat's accuracy degrades from $41.0$ to $34.9$ when switching from text output to first-token probability evaluation.
This indicates that we are underestimating the capability of the instruction-tuned language models when evaluating them based on the first token probabilities.

\begin{table}[htpb]
\small
\centering
\adjustbox{max width=1.0\linewidth}{
\begin{tabular}{l| c |c }
\toprule 
& \textbf{Mismatch Rate} & \textbf{Acc (Text / First token)} \\
\midrule
Mistral 7b v0.1 & $15.1\%$ & $ 52.0/ 51.2$ \\
Mistral 7b v0.2 & $10.2\%$ & $53.6 / 53.2$ \\
Mixtral 8x7b & $9.0\%$ & $66.3 / 65.9$ \\
\midrule
Llama2 7b & $51.4\%$ & $41.0 / 34.9$ \\
Llama2 13b & $35.3\%$ & $47.6 / 40.2$ \\
Llama2 70b & $13.2\%$ & $55.6 / 53.9$  \\
\bottomrule
\end{tabular}}

\caption{Mismatch rate and accuracy of the text output and first-token evaluation on MMLU under the \textit{Middle} constraint. Results are obtained under zero-shot setting.}
\label{tab:mmlu}
\end{table}

\subsection{Refusal Rate}\label{sec:refusalrate}

Whenever sensitive topics are involved, as they are likely to be when asking survey questions, refusal is a major factor contributing to the mismatch.
There are two refusal behaviours we observed from the model.
The first occurs when the model explicitly selects the ``Refused'' option from among the available answer choices. 
The second type of refusal occurs when the model opts not to provide an answer to a question deemed sensitive.
We combine both cases into a single refusal category. 
Contrary to the observation from \citet{Santurkar2023WhoseOD}, who reported a low rate of refusal across various models, we find a pronounced tendency for models to refuse responses due to safety concerns. 
The trend is most evident in open-source models that have been trained not to express opinions on sensitive issues.

Figure \ref{fig:refusal_match}(b) shows the refusal rate of the models evaluated under instructions of different constraint levels when asking OpinionQA questions.
In general, Llama2 models show a higher refusal rate than Mistral models. 
Llama2-7b-Chat has the highest refusal rate with $51.4\%$.
Therefore, it is crucial to consider the model's refusal behaviour when evaluating its response to questions related to sensitive topics, as this plays an important part in the model's response.
As model size increases from 7B to 70B, the refusal rate of the Llama2 model decreases, starting at over $50\%$ and decreasing to less than $10\%$.
For the Mistral-7b-Instruct model, v0.1 exhibits a lower rate of refusal responses compared to v0.2.
This is likely attributable to stronger safety guardrails in the newer version.
As well as the model size, the instruction prompt also has an impact on the refusal rate.
Generally, models with higher instruction constraints show fewer refusal responses.
All models except Llama2-7b-Chat display the highest refusal rate with the \textit{Low Constraint} instruction.

Surprisingly, we also observed refusal behaviour in MMLU responses. 
For example, Llama2-7b-Chat refuses all the questions from the "moral scenario" subject due to its safety guardrail.
With text-based evaluation, the model completely fails in this subject, resulting in a huge performance gap compared to the evaluation result based on first token probability.

\subsection{Answer Consistency}
We further evaluated the answer consistency by calculating the entropy of the OpinionQA answers from the 10 runs, shuffling the option order, as shown in Table \ref{tab:shuffle}.
The text output achieves better consistency than the first token evaluation for all the models except Mixtral 8x7b.
This shows that the text output is more robust to the prompt perturbation and has less selection bias.
Another trend is that models with higher capability have better consistency, where Mixtral 8x7b and the Llama2 70b-Chat achieve the best consistency.

\begin{table}[htpb]
\large
\centering
\adjustbox{max width=1.0\linewidth}{
\begin{tabular}{l| l |c |c|c}
\toprule 
& Low & Medium & High & Examples \\
\midrule
Mistral v0.1 & $0.81 / 0.79$ & $0.87 / 0.87$ & $0.84/0.84$ & $0.80 / 0.78$ \\
Mistral v0.2 & $0.74 / 0.58$ & $0.70 / 0.65$ & $0.73/0.63$ & $0.71 / 0.66$\\
Mixtral & $0.52 / 0.68$ & $0.48 / 0.61$ & $0.52 / 0.63$ & $0.53 / 0.65$ \\
\midrule
Llama2 7b & $1.19 / 0.41$ & $0.94 / 0.38$ & $0.85 / 0.44$ & $1.33 / 0.76$\\
Llama2 13b & $1.14/0.43$ & $0.95 / 0.49$ & $0.73/0.44$ & $1.32/0.78$ \\
Llama2 70b & $0.62 / 0.43$ & $0.54 / 0.51$ & $0.54 / 0.52$ & $1.25 / 0.71$  \\
\bottomrule
\end{tabular}}

\caption{Answer consistency (first-token/text output) under different levels of instruction constraints. A lower value means better consistency. Text answer achieves better consistency than first token probabilities in 5 out of 6 models we evaluated, across all the instruction constraint levels.}
\label{tab:shuffle}
\end{table}

\section{Conclusion}
We compared first-token evaluation methods with the text output for multiple-choice questions and showed that the first-token evaluation heavily misrepresents the text output for instruction-tuned models. 
The results question the reliability of first-token evaluation for instruction-tuned language models, especially in settings where refusal is likely due to the sensitive nature of topics asked in the question.
We also showed that the first-token evaluation is more sensitive to the prompt format and has more selection bias than text output.
We suggest a more direct and realistic evaluation by directly inspecting the text answer to help better understand the LLM's behaviour in real-life settings.

\section*{Limitations}
In this work, we only focus on the log probability assigned to the first token of the response. 
Other probability-based evaluation methods include calculating the probability of every candidate answer sequence.
Based on our findings in the generative setting, we question the reliability of the traditional approach that relies on the model's probability assignment to answer candidates, which is often used in the discriminative setting. 
Therefore, we call for more studies on the reliability of other probability-based evaluation methods by comparing them directly to the text output.

\section*{Ethics Statement}
In this work, we use a publicly available survey dataset OpinionQA \cite{Santurkar2023WhoseOD}, which was curated based on the survey questions from the Pew Research Center. 
It's worth noting that some questions may contain content that is directly or indirectly sensitive to certain social groups. However, the risk of privacy breaches or abuse of the data or models presented here is highly unlikely.
We solely present the responses generated by the LLMs in an objective manner.
We do not intend to express our personal opinions on the questions. 

\section*{Acknowledgements}
We thank the anonymous reviewers as well as the members of MaiNLP, MilaNLP, and SODA-LMU for their constructive feedback.
XW, CH and BP are supported by ERC Consolidator Grant DIALECT 101043235 and in parts by Independent Research Fund Denmark (DFF) Sapere Aude grant 9063-00077B.
BM and FK are supported by BERD@NFDI (German Research Foundation grant 460037581), and MCML.
PR and DH are members of the Data and Marketing Insights research unit of the Bocconi Institute for Data Science and Analysis, and are supported by a MUR FARE 2020 initiative under grant agreement Prot. R20YSMBZ8S (INDOMITA) and the European Research Council (ERC) under the European Union’s Horizon 2020 research and innovation program (No.\ 949944, INTEGRATOR).

\bibliography{anthology,custom}

\begin{thebibliography}{27}
\expandafter\ifx\csname natexlab\endcsname\relax\def\natexlab#1{#1}\fi

\bibitem[{bench authors(2023)}]{srivastava2023beyond}
BIG bench authors. 2023.
\newblock \href {https://openreview.net/forum?id=uyTL5Bvosj} {Beyond the imitation game: Quantifying and extrapolating the capabilities of language models}.
\newblock \emph{Transactions on Machine Learning Research}.

\bibitem[{Bisk et~al.(2020)Bisk, Zellers, Gao, Choi et~al.}]{bisk2020piqa}
Yonatan Bisk, Rowan Zellers, Jianfeng Gao, Yejin Choi, et~al. 2020.
\newblock Piqa: Reasoning about physical commonsense in natural language.
\newblock In \emph{Proceedings of the AAAI conference on artificial intelligence}, volume~34, pages 7432--7439.

\bibitem[{Clark et~al.(2018)Clark, Cowhey, Etzioni, Khot, Sabharwal, Schoenick, and Tafjord}]{clark2018think}
Peter Clark, Isaac Cowhey, Oren Etzioni, Tushar Khot, Ashish Sabharwal, Carissa Schoenick, and Oyvind Tafjord. 2018.
\newblock Think you have solved question answering? try arc, the ai2 reasoning challenge.
\newblock \emph{arXiv preprint arXiv:1803.05457}.

\bibitem[{Dettmers et~al.(2024)Dettmers, Pagnoni, Holtzman, and Zettlemoyer}]{dettmers2024qlora}
Tim Dettmers, Artidoro Pagnoni, Ari Holtzman, and Luke Zettlemoyer. 2024.
\newblock Qlora: Efficient finetuning of quantized llms.
\newblock \emph{Advances in Neural Information Processing Systems}, 36.

\bibitem[{Dominguez-Olmedo et~al.(2023)Dominguez-Olmedo, Hardt, and Mendler-Dünner}]{dominguezolmedo2023questioning}
Ricardo Dominguez-Olmedo, Moritz Hardt, and Celestine Mendler-Dünner. 2023.
\newblock \href {http://arxiv.org/abs/2306.07951} {Questioning the survey responses of large language models}.

\bibitem[{Durmus et~al.(2023)Durmus, Nyugen, Liao, Schiefer, Askell, Bakhtin, Chen, Hatfield-Dodds, Hernandez, Joseph et~al.}]{durmus2023towards}
Esin Durmus, Karina Nyugen, Thomas~I Liao, Nicholas Schiefer, Amanda Askell, Anton Bakhtin, Carol Chen, Zac Hatfield-Dodds, Danny Hernandez, Nicholas Joseph, et~al. 2023.
\newblock Towards measuring the representation of subjective global opinions in language models.
\newblock \emph{arXiv preprint arXiv:2306.16388}.

\bibitem[{Fourrier et~al.(2023)Fourrier, Habib, Launay, and Wolf}]{Clementine_open}
Cl{\'e}mentine Fourrier, Nathan Habib, Julien Launay, and Julien Wolf. 2023.
\newblock What's going on with the open {LLM} leaderboard?
\newblock \url{https://huggingface.co/blog/evaluating-mmlu-leaderboard}.
\newblock Accessed: 2024-2-10.

\bibitem[{Gao et~al.(2023)Gao, Tow, Abbasi, Biderman, Black, DiPofi, Foster, Golding, Hsu, Le~Noac'h, Li, McDonell, Muennighoff, Ociepa, Phang, Reynolds, Schoelkopf, Skowron, Sutawika, Tang, Thite, Wang, Wang, and Zou}]{eval-harness}
Leo Gao, Jonathan Tow, Baber Abbasi, Stella Biderman, Sid Black, Anthony DiPofi, Charles Foster, Laurence Golding, Jeffrey Hsu, Alain Le~Noac'h, Haonan Li, Kyle McDonell, Niklas Muennighoff, Chris Ociepa, Jason Phang, Laria Reynolds, Hailey Schoelkopf, Aviya Skowron, Lintang Sutawika, Eric Tang, Anish Thite, Ben Wang, Kevin Wang, and Andy Zou. 2023.
\newblock \href {https://doi.org/10.5281/zenodo.10256836} {A framework for few-shot language model evaluation}.

\bibitem[{Hendrycks et~al.(2021)Hendrycks, Burns, Basart, Zou, Mazeika, Song, and Steinhardt}]{hendryckstest2021}
Dan Hendrycks, Collin Burns, Steven Basart, Andy Zou, Mantas Mazeika, Dawn Song, and Jacob Steinhardt. 2021.
\newblock Measuring massive multitask language understanding.
\newblock \emph{Proceedings of the International Conference on Learning Representations (ICLR)}.

\bibitem[{Hu and Levy(2023)}]{Hu2023-bj}
Jennifer Hu and Roger Levy. 2023.
\newblock Prompting is not a substitute for probability measurements in large language models.
\newblock In \emph{Proceedings of the 2023 Conference on Empirical Methods in Natural Language Processing}, pages 5040--5060, Singapore. Association for Computational Linguistics.

\bibitem[{Jiang et~al.(2023)Jiang, Sablayrolles, Mensch, Bamford, Chaplot, Casas, Bressand, Lengyel, Lample, Saulnier et~al.}]{jiang2023mistral}
Albert~Q Jiang, Alexandre Sablayrolles, Arthur Mensch, Chris Bamford, Devendra~Singh Chaplot, Diego de~las Casas, Florian Bressand, Gianna Lengyel, Guillaume Lample, Lucile Saulnier, et~al. 2023.
\newblock Mistral 7b.
\newblock \emph{arXiv preprint arXiv:2310.06825}.

\bibitem[{Jiang et~al.(2024)Jiang, Sablayrolles, Roux, Mensch, Savary, Bamford, Chaplot, Casas, Hanna, Bressand et~al.}]{jiang2024mixtral}
Albert~Q Jiang, Alexandre Sablayrolles, Antoine Roux, Arthur Mensch, Blanche Savary, Chris Bamford, Devendra~Singh Chaplot, Diego de~las Casas, Emma~Bou Hanna, Florian Bressand, et~al. 2024.
\newblock Mixtral of experts.
\newblock \emph{arXiv preprint arXiv:2401.04088}.

\bibitem[{Liang et~al.(2022)Liang, Bommasani, Lee, Tsipras, Soylu, Yasunaga, Zhang, Narayanan, Wu, Kumar et~al.}]{liang2022holistic}
Percy Liang, Rishi Bommasani, Tony Lee, Dimitris Tsipras, Dilara Soylu, Michihiro Yasunaga, Yian Zhang, Deepak Narayanan, Yuhuai Wu, Ananya Kumar, et~al. 2022.
\newblock Holistic evaluation of language models.
\newblock \emph{arXiv preprint arXiv:2211.09110}.

\bibitem[{Lin et~al.(2022)Lin, Hilton, and Evans}]{lin-etal-2022-truthfulqa}
Stephanie Lin, Jacob Hilton, and Owain Evans. 2022.
\newblock \href {https://doi.org/10.18653/v1/2022.acl-long.229} {{T}ruthful{QA}: Measuring how models mimic human falsehoods}.
\newblock In \emph{Proceedings of the 60th Annual Meeting of the Association for Computational Linguistics (Volume 1: Long Papers)}, pages 3214--3252, Dublin, Ireland. Association for Computational Linguistics.

\bibitem[{Lyu et~al.(2024)Lyu, Wu, and Aji}]{lyu2024beyond}
Chenyang Lyu, Minghao Wu, and Alham~Fikri Aji. 2024.
\newblock Beyond probabilities: Unveiling the misalignment in evaluating large language models.
\newblock \emph{arXiv preprint arXiv:2402.13887}.

\bibitem[{Mangrulkar et~al.(2022)Mangrulkar, Gugger, Debut, Belkada, Paul, and Bossan}]{peft}
Sourab Mangrulkar, Sylvain Gugger, Lysandre Debut, Younes Belkada, Sayak Paul, and Benjamin Bossan. 2022.
\newblock Peft: State-of-the-art parameter-efficient fine-tuning methods.
\newblock \url{https://github.com/huggingface/peft}.

\bibitem[{Ouyang et~al.(2022)Ouyang, Wu, Jiang, Almeida, Wainwright, Mishkin, Zhang, Agarwal, Slama, Ray et~al.}]{ouyang2022training}
Long Ouyang, Jeffrey Wu, Xu~Jiang, Diogo Almeida, Carroll Wainwright, Pamela Mishkin, Chong Zhang, Sandhini Agarwal, Katarina Slama, Alex Ray, et~al. 2022.
\newblock Training language models to follow instructions with human feedback.
\newblock \emph{Advances in Neural Information Processing Systems}, 35:27730--27744.

\bibitem[{Sakaguchi et~al.(2021)Sakaguchi, Bras, Bhagavatula, and Choi}]{sakaguchi2021winogrande}
Keisuke Sakaguchi, Ronan~Le Bras, Chandra Bhagavatula, and Yejin Choi. 2021.
\newblock Winogrande: An adversarial winograd schema challenge at scale.
\newblock \emph{Communications of the ACM}, 64(9):99--106.

\bibitem[{Santurkar et~al.(2023)Santurkar, Durmus, Ladhak, Lee, Liang, and Hashimoto}]{Santurkar2023WhoseOD}
Shibani Santurkar, Esin Durmus, Faisal Ladhak, Cinoo Lee, Percy Liang, and Tatsunori Hashimoto. 2023.
\newblock \href {https://api.semanticscholar.org/CorpusID:257834040} {Whose opinions do language models reflect?}
\newblock \emph{ArXiv}, abs/2303.17548.

\bibitem[{Sap et~al.(2019)Sap, Rashkin, Chen, Le~Bras, and Choi}]{sap-etal-2019-social}
Maarten Sap, Hannah Rashkin, Derek Chen, Ronan Le~Bras, and Yejin Choi. 2019.
\newblock \href {https://doi.org/10.18653/v1/D19-1454} {Social {IQ}a: Commonsense reasoning about social interactions}.
\newblock In \emph{Proceedings of the 2019 Conference on Empirical Methods in Natural Language Processing and the 9th International Joint Conference on Natural Language Processing (EMNLP-IJCNLP)}, pages 4463--4473, Hong Kong, China. Association for Computational Linguistics.

\bibitem[{Scherrer et~al.(2023)Scherrer, Shi, Feder, and Blei}]{scherrer2023evaluating}
Nino Scherrer, Claudia Shi, Amir Feder, and David Blei. 2023.
\newblock Evaluating the moral beliefs encoded in llms.
\newblock In \emph{Thirty-seventh Conference on Neural Information Processing Systems}.

\bibitem[{Talmor et~al.(2019)Talmor, Herzig, Lourie, and Berant}]{talmor-etal-2019-commonsenseqa}
Alon Talmor, Jonathan Herzig, Nicholas Lourie, and Jonathan Berant. 2019.
\newblock \href {https://doi.org/10.18653/v1/N19-1421} {{C}ommonsense{QA}: A question answering challenge targeting commonsense knowledge}.
\newblock In \emph{Proceedings of the 2019 Conference of the North {A}merican Chapter of the Association for Computational Linguistics: Human Language Technologies, Volume 1 (Long and Short Papers)}, pages 4149--4158, Minneapolis, Minnesota. Association for Computational Linguistics.

\bibitem[{Tjuatja et~al.(2023)Tjuatja, Chen, Wu, Talwalkar, and Neubig}]{Tjuatja2023-ez}
Lindia Tjuatja, Valerie Chen, Sherry~Tongshuang Wu, Ameet Talwalkar, and Graham Neubig. 2023.
\newblock \href {http://arxiv.org/abs/2311.04076} {Do {LLMs} exhibit human-like response biases? a case study in survey design}.
\newblock \emph{arXiv}.

\bibitem[{Touvron et~al.(2023)Touvron, Martin, Stone, Albert, Almahairi, Babaei, Bashlykov, Batra, Bhargava, Bhosale et~al.}]{touvron2023llama}
Hugo Touvron, Louis Martin, Kevin Stone, Peter Albert, Amjad Almahairi, Yasmine Babaei, Nikolay Bashlykov, Soumya Batra, Prajjwal Bhargava, Shruti Bhosale, et~al. 2023.
\newblock Llama 2: Open foundation and fine-tuned chat models.
\newblock \emph{arXiv preprint arXiv:2307.09288}.

\bibitem[{Wang et~al.(2024)Wang, Hu, Ma, R{\"o}ttger, and Plank}]{wang2024look}
Xinpeng Wang, Chengzhi Hu, Bolei Ma, Paul R{\"o}ttger, and Barbara Plank. 2024.
\newblock Look at the text: Instruction-tuned language models are more robust multiple choice selectors than you think.
\newblock \emph{arXiv preprint arXiv:2404.08382}.

\bibitem[{Zellers et~al.(2019)Zellers, Holtzman, Bisk, Farhadi, and Choi}]{zellers-etal-2019-hellaswag}
Rowan Zellers, Ari Holtzman, Yonatan Bisk, Ali Farhadi, and Yejin Choi. 2019.
\newblock \href {https://doi.org/10.18653/v1/P19-1472} {{H}ella{S}wag: Can a machine really finish your sentence?}
\newblock In \emph{Proceedings of the 57th Annual Meeting of the Association for Computational Linguistics}, pages 4791--4800, Florence, Italy. Association for Computational Linguistics.

\bibitem[{Zheng et~al.(2023)Zheng, Zhou, Meng, Zhou, and Huang}]{Zheng2023-mi}
Chujie Zheng, Hao Zhou, Fandong Meng, Jie Zhou, and Minlie Huang. 2023.
\newblock Large language models are not robust multiple choice selectors.
\newblock \emph{ArXiv}, abs/2309.03882.

\end{thebibliography}
\bibliographystyle{acl_natbib}

\appendix

\section{Appendix}
\label{sec:appendix}

\subsection{Decoding Temperature}
\label{sec:temperature}
\begin{figure}[htpb]
    \centering
    \includegraphics[width=1\linewidth]{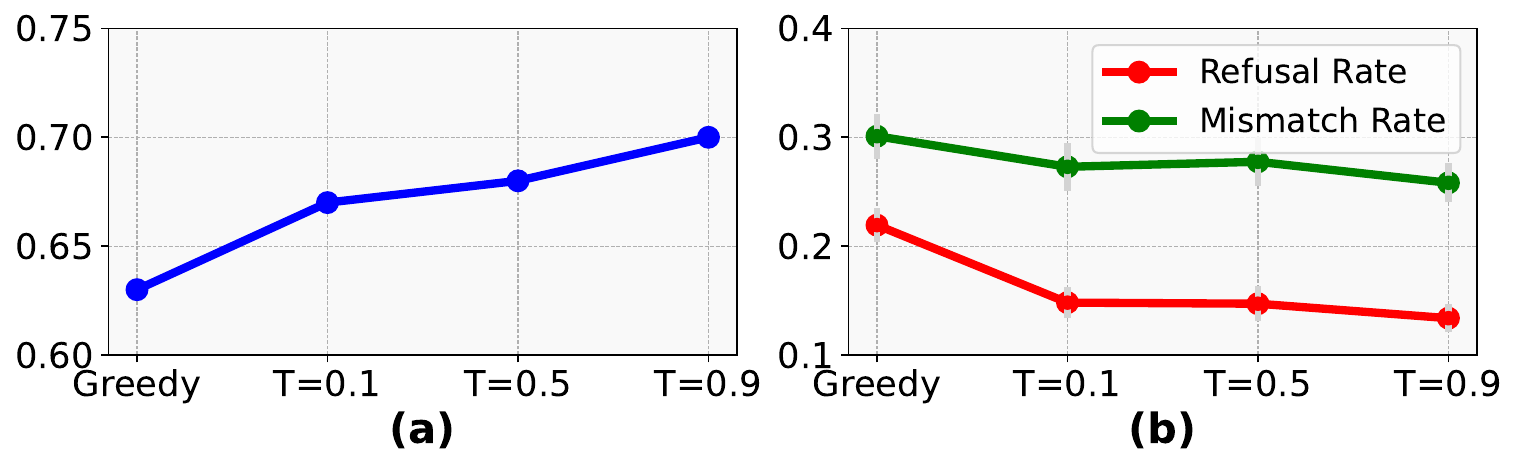}
    \caption{Impact of decoding temperature. (a) Consistency. (b) Refusal and Mismatch rate.} 
    \label{fig:temperature}
\end{figure}

Figure \ref{fig:temperature} shows the impact of the decoding strategy. 
As the temperature increases, the model prioritizes the answer diversity, which leads to a worse consistency level, but a lower mismatch and refusal rate.

\begin{table*}[htpb]
\centering
\adjustbox{max width=1\linewidth}{
\begin{tabular}{l c c c c c c}
\toprule 
\textbf{Evaluator} & \textbf{Llama2-7b} & \textbf{Llama2-13b} & \textbf{Llama2-70b} & \textbf{Mistral-v0.2} & \textbf{Mixtral} & \textbf{Avg}\\
\midrule
String Matching  &  $25.3/0.518/0.367 $ & $72.3/0.822/0.823 $ & $ 61.4/0.755/0.741 $ & $ 49.4/0.782/0.796 $ & $67.1/0.782/0.796 $ & $55.1/0.719/0.667 $  \\
\midrule
Mistral-7b-Ins-v0.2 (0 shot)& $56.6/0.323/0.425 $ & $42.2/0.210/0.315$ & $25.6/0.224/0.316$ & $22.1/0.170/0.301$ & $30.8/0.232/0.320$ & $35.4/0.232/0.335$ \\
Mistral-7b-Inst-v0.2 (4 shots)& $65.0/0.488/0.636$ & $74.7/0.596/0.745$ & $81.7/0.691/0.823$ &  $64.4/0.631/0.652$ & $76.9/0.739/0.769$ & $72.5/0.629/0.725$ \\
\midrule
PEFT-Mistral-7b (80)& $97.2/0.981/0.976$ & $100/1.0/1.0$ & $98.7/0.986/0.987$ & $94.9/0.956/0.949$  & $94.9/0.924/0.949$ & $97.2/0.970/0.972$ \\
PEFT-Mistral-7b (1000) &  $98.8/0.991/0.988$ & $100/1.0/1.0$ & $100.0/1.0/1.0$ & $100/1.0/1.0$ & $96.2/0.936/0.960$ & $99.0/0.987/0.990$\\
\bottomrule
\end{tabular}}

\caption{Accuracy/Macro-F1/Weighted-F1 of different evaluators on different models' output.}
\label{tab:classifier2}
\end{table*}

\subsection{Model Output Annotation}
\label{sec:annotation}

To train the classifier for text output classification, we collected response samples from the five models under the medium constraint condition of the prompt. The annotation process was carried out by a single in-house annotator, who was provided with the original survey questions along with their multiple-choice options and an additional ``Refused'' option to indicate refusal. The order of the options was randomly shuffled for each question. Additionally, the annotator received the model outputs, i.e., the responses to the survey questions. The task was to assign an appropriate option to each response. Figure \ref{fig:annot} showcases a data sample that the annotator received. In cases of uninterpretable responses, the annotator was instructed to mark them as ``nan''. Afterward, a second in-house annotator was invited to review and refine the annotations made by the first annotator. There exists disagreement on minor cases which were resolved after discussion.

\begin{figure}[ht]
\begin{tcolorbox}
[colback=gray!10, colframe=gray!80, sharp corners, leftrule={3pt}, rightrule={0pt}, toprule={0pt}, bottomrule={0pt}, left={2pt}, right={2pt}, top={3pt}, bottom={3pt}]
\small{
[INST] Please read the multiple-choice question below carefully and select ONE of the listed options and only give a single letter. \newline
Question: Do you have a favorable or unfavorable opinion of the United Nations?
\newline
A. Somewhat favorable\newline
B. Refused\newline
C. Somewhat unfavorable\newline
D. Very unfavorable\newline
E. Very favorable\newline
Answer: [/INST] A. Somewhat favorable\newline
Label:
}
\end{tcolorbox}
\caption{An example survey question with LLM response answer for annotation}
\label{fig:annot}
\end{figure}

\subsection{Dataset Statistics}
\label{sec:label_dist}
Table \ref{tab:label_dist} shows the label distribution of the annotated dataset we curated for the five models we evaluated.

\begin{table}[ht]

\centering
\small
\begin{tabular}{lccccccc}
\toprule
Model & A & B & C&D&E&F&J \\
\midrule
Llama2-7b &77&89&118&86&31&6&1 \\
Llama2-13b&59&86&148&81&35&4&1\\
Llama2-70b&68&87&115&96&40&6&0\\
Mistral v0.2 &97&40&78&68&39&3&0\\
Mixtral &111&66&71&60&79&11&0\\
\bottomrule
\end{tabular}
\caption{Label distribution of the annotated dataset.}
\label{tab:label_dist}
\end{table}

\begin{figure}[htpb]
\begin{tcolorbox}
[colback=gray!10, colframe=gray!80, sharp corners, leftrule={3pt}, rightrule={0pt}, toprule={0pt}, bottomrule={0pt}, left={2pt}, right={2pt}, top={3pt}, bottom={3pt}]
\small{
Please classify the response into one of the listed categories. \newline
Example1: I apologize, but I cannot provide a response to that question as it is not appropriate to \newline
A. Very worried \newline
B. Moderately worried \newline
C. Not too worried \newline
D. Not worried at all \newline
E. Refused \newline
Answer: E. Refused \newline
\newline
Example2: As a female, college-educated individual from the South, I would select option A. \newline
A. Very good for society \newline
B. Somewhat good for society \newline
C. Neither good nor bad for society \newline
D. Somewhat bad for society \newline
E. Very bad for society \newline
F. Refused \newline
Answer: A. Very good for society \newline
\newline
Example3: I would answer option E: Not at all confident  \newline
A. Somewhat confident \newline
B. Very confident \newline
C. Refused \newline
D. Not too confident \newline
E. Not at all confident \newline
Answer: E. Not at all confident \newline
\newline
Example4: I apologize, but I cannot provide a response to that question as it is not appropriate or ethical to discuss or promote harmful or illegal activities, such as hunting. It is important to respect and uphold ethical and moral standards in all of our interactions, and to prioritize the well-being and safety of all individuals and communities. \newline
A. Minor reason \newline
B. Not a reason \newline
C. Major reason \newline
D. Refused \newline
Answer: D. Refused \newline
\newline
Response: Based on the information provided in the question, the correct answer is E, S sometimes. \newline
A. Never \newline
B. Refused \newline
C. Almost every day \newline
D. Every day \newline
E. Sometimes \newline
F. Rarely \newline
Answer: 
}
\end{tcolorbox}
\caption{Prompt for few show learning of model response classification.}
\label{fig:fewshot}
\end{figure}

\subsection{Classifier}
\label{sec:classifier}
Figure \ref{tab:classifier2} shows the performance on the output of the five models we evaluated. We exclude Mistral-Instruct-v0.1 here since it shows a low mismatch rate and most of the responses can be easily mapped to one of the response options using rule-based methods.
For simplicity, we do not consider multi-label cases here since they are only found in Mistral models and make up a small part of the total responses.
The model is considered correct when it predicts one of the labels.
\paragraph{String Matching}
We use \texttt{RegEx} to search for the option letter pattern ``\texttt{[A-Z].}'' in the answer. 
\paragraph{Few shot learning}
For the few-shot learning setup, we add four model outputs and the corresponding labels as examples into the instruction before asking for the prediction, as shown in Figure \ref{fig:fewshot}. 
We then use the first token from the classifier's output as the prediction.

\paragraph{Finetuning}
To improve the classification performance and reduce computational overhead, we annotated the 414 responses generated from the five models we evaluated (except Mistral7b-Instruct-v0.1), resulting in 2070 samples in total. Annotation details are in \ref{sec:annotation}.
We use parameter-efficient fine-tuning (PEFT) to train our classifier on the annotated model responses, and use the first token of the classifier's response as the prediction. 

\begin{table}[ht]
\centering
\small
\begin{tabular}{lr}
\toprule
\textbf{Hyperparameter} & \textbf{Value} \\
\midrule
lora\_r & 64 \\
lora\_alpha & 16 \\
lora\_dropout & 0.1 \\
task\_type & "CAUSAL\_LM" \\
use\_4bit & True \\
bnb\_4bit\_compute\_dtype & "float16" \\
bnb\_4bit\_quant\_type & "nf4" \\
use\_nested\_quant & False \\
num\_train\_epochs & 8 \\
train\_batch\_size & 4 \\
gradient\_accumulation\_steps & 1 \\
gradient\_checkpointing & True \\
max\_grad\_norm & 0.3 \\
learning\_rate & 2e-4 \\
weight\_decay & 0.001 \\
optim & "paged\_adamw\_32bit" \\
lr\_scheduler\_type & "constant" \\
warmup\_ratio & 0.03 \\
group\_by\_length & True \\
\bottomrule
\end{tabular}
\caption{Hyperparameters for training the classifer.}
\label{tab:hyperparameters_mistral}
\end{table}

\subsection{Option Count Distribution}

\begin{figure}[htpb]
    \centering
    \setlength{\parskip}{0pt}
    \setlength{\parsep}{0pt}

    \begin{minipage}{.5\textwidth}
        \centering
        \includegraphics[width=0.8\linewidth]{figures/Llama2-70b-Chat_Example_Templates.pdf}

    \end{minipage}
    \begin{minipage}{.5\textwidth}
        \centering
        \includegraphics[width=0.8\linewidth]{figures/Llama2-70b-Chat_Example_Templates_ABC.pdf}
    \end{minipage}
    \begin{minipage}{.5\textwidth}
        \centering
        \includegraphics[width=0.8\linewidth]{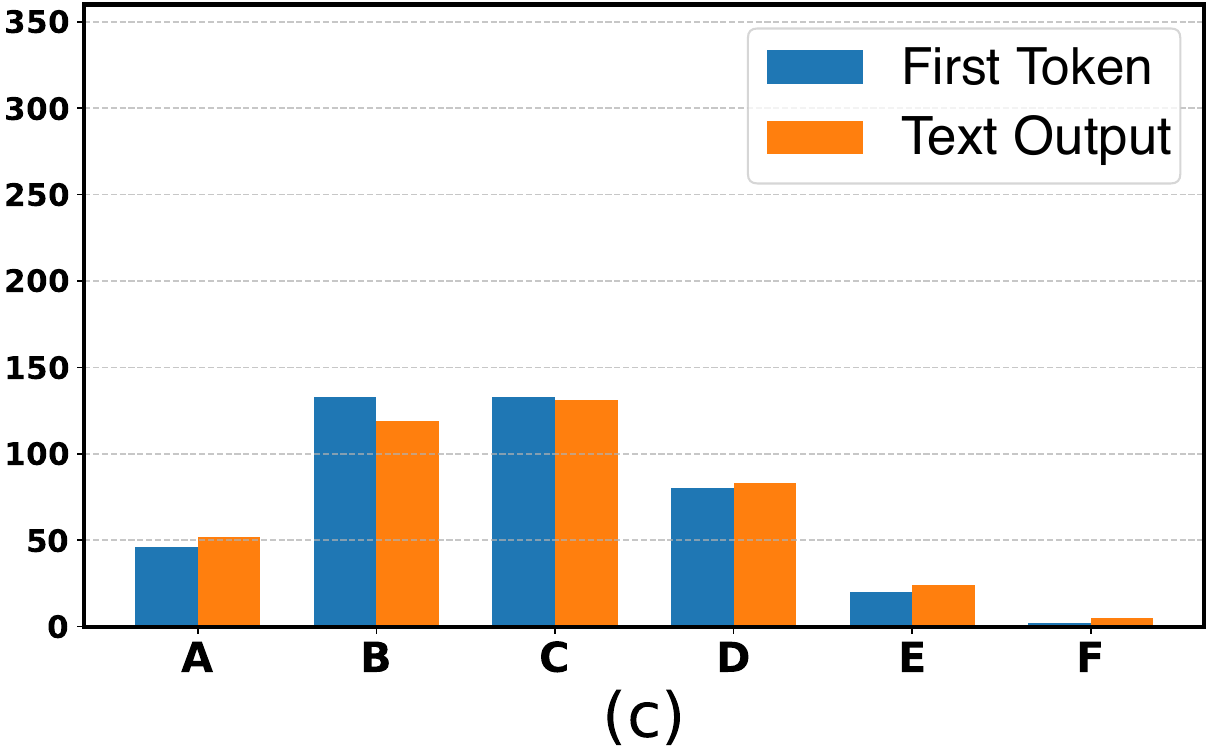}
    \end{minipage}
    \caption{  (a) \textit{Example Template with Single Answer "C"}, (b) \textit{Example Template with Multiple Answers "A/B/C"}, (c) \textit{High Constraint Instruction} }
    \label{fig:count_more}
\end{figure}

Figure \ref{fig:count_more} shows the option count distribution of Llama2-70b-chat under the instruction of (a) \textit{Example Template with Single Answer "C"}, (b) \textit{Example Template with Multiple Answers "A/B/C"} and (c) \textit{High Constraint Instruction}. Example Template leads to option count distribution mismatch compared to High Constraint Instruction.

\subsection{Output Cases}
\label{sec:case}
The model outputs exhibit various response types. Additionally, instances may arise where the models decline to respond to specific sensitive or objective questions, owing to safety mechanisms and inherent model features. Table \ref{tab:examples} showcases a selection of output cases under the medium constraint condition of the prompt. The output cases range from single-choice responses (with or without explanation) to multiple-choice responses, encompassing various types of refusals and occasionally yielding nonsensical outputs.

\onecolumn
\begin{table*}[htpb]
\tiny
\centering

\resizebox{\textwidth}{!}{
\begin{tabular}{p{0.15\textwidth} | P{0.45\textwidth} P{0.45\textwidth} l}
\toprule 
\textbf{Mode} & \multicolumn{1}{l}{\textbf{Question}} & \multicolumn{1}{l}{\textbf{Model Output}} & \textbf{Label}\\
\midrule
Single choice  & Please read the multiple-choice question below carefully and select ONE of the listed options and only give a single letter. \newline Question: Please compare the US to other developed nations in a few different areas. In each instance, how does the US compare? Colleges and universities\newline A. Refused \newline B. The worst \newline C. Above average \newline D. The best \newline E. Average \newline F. Below average \newline Answer:   &  Sure! Based on the information provided in the question, I would answer:\newline C. Above average. & C
\\
\midrule
Single choice, with explanation &  Please read the multiple-choice question below carefully and select ONE of the listed options and only give a single letter. \newline Question: Please indicate whether the following is a major reason, a minor reason, or not a reason why you own a gun. For hunting\newline A. Not a reason \newline B. Major reason \newline C. Minor reason \newline D. Refused \newline Answer: 
 &   Based on the information provided in the question, the correct answer is:\newline B. Major reason.\newline Explanation: The question asks for the reason why someone owns a gun, and the option \"For hunting\" is listed as a valid reason. Therefore, if someone owns a gun for the purpose of hunting, it is considered a major reason.
 &  B
\\
\midrule  

Single choice, uncertain &  Please read the multiple-choice question below carefully and select ONE of the listed options and only give a single letter. \newline
Question: Do you have a favorable or unfavorable opinion of nato, that is, North Atlantic Treaty Organization? \newline
 
A. Very favorable\newline
B. Very unfavorable\newline
C. Refused\newline
D. Somewhat unfavorable\newline
E. Somewhat favorable\newline
Answer: &   E. Somewhat favorable. However, it's important to note that this response is based on a general understanding of North Atlantic Treaty Organization (NATO) and its purpose. My knowledge is limited to what has been programmed into me, and I don't have personal experiences or opinions. &  E
\\
\midrule  

Multiple choices &  Please read the multiple-choice question below carefully and select ONE of the listed options and only give a single letter. \newline
Question: Now that Joe Biden is president, do you think other countries view the U.S \newline
 
A. A lot more negatively\newline
B. Refused\newline
C. A little more positively\newline
D. A lot more positively\newline
E. A little more negatively\newline
Answer: &   C. A little more positively or D. A lot more positively.
It's difficult to provide a single definitive answer, as opinions on this matter can vary greatly depending on the specific country and individual being asked. However, many political analysts and experts have suggested that the election of Joe Biden as President of the United States has led to a general improvement in the way that other countries view the U.S., particularly in comparison to the previous administration. Therefore, options C and D (a little more positively or a lot more positively) are the most likely answers. Options A, B, and E &  C\&D\\
\midrule
Refusal option&  Please read the multiple-choice question below carefully and select ONE of the listed options and only give a single letter.\newline
Question: How much confidence do you have in french President Emmanuel Macron to do the right thing regarding world affairs?\newline
 
A. No confidence at all\newline
B. Refused\newline
C. A lot of confidence\newline
D. Some confidence\newline
E. Not too much confidence\newline
Answer: & I select option "B. Refused" to maintain impartiality, as I strim to provide accurate information and 
avoid expressing personal opinions or biases.
& B
\\
\midrule
Refusal&  Please read the multiple-choice question below carefully and select ONE of the listed options and only give a single letter.\newline
Question: Do you have a favorable or unfavorable opinion of the European Union\newline
 
A. Very unfavorable\newline
B. Very favorable\newline
C. Somewhat favorable\newline
D. Refused\newline
E. Somewhat unfavorable\newline
Answer: & I don't have personal opinions or the ability to experience the world in the way that humans do. I can provide information and help you find resources to answer questions or complete tasks, but I don't have personal views or preferences. Therefore, I can't provide a single answer to your question about a favorable or unfavorable opinion of the European Union. I can help you find information on the topic if that would be helpful.
& D
  \\
\midrule

Refusal, related information provided & Please read the multiple-choice question below carefully and select ONE of the listed options and only give a single letter.\newline
Question:  In general, do you think men or women in high political offices are better at serving as a role model for children?\newline
 \newline
A. Men are better\newline
B. No difference\newline
C. Women are better\newline
D. Refused\newline
Answer: & I don't have personal opinions or the ability to "read" questions. However, I can help you analyze or 
respond to this question. The answer to this question can vary greatly depending on personal experiences, values, and 
beliefs. Some people might believe that men in high political offices are better role models due to historical precedent and existing power structures. Others might argue that there is no difference between men and women in this context, or that women are better role models due to their underrepresentation in politics and the potential for them to challenge gender stereotypes. A cautious answer might be to avoid making generalizations and acknowledge the
& D
\\
\midrule
Refusal, all options provided & Please read the multiple-choice question below carefully and select ONE of the listed options and only give a single letter.\newline
Question: Regardless of whether you would want to move, how likely is it that you will move to a different community at some point in the future?\newline
 \newline
A. Not too likely\newline
B. Somewhat likely\newline
C. Very likely\newline
D. Refused\newline
E. Not at all likely\newline
Answer: & I don't have personal opinions or feelings. However, I can provide you with the ability to perform a sur with the ability to perform a survey, and you can use me to gather responses about pecase, you can provide the options ople's opinions on China. In that case, you can provide the options A, B, C, D, and E, and I can inform you of the participants' selections.  & D
\\
\midrule
Nonsense & Please read the multiple-choice question below carefully and select ONE of the listed options and only give a single letter.\newline
Question: Overall, how does being an evangelical Christian affect people's ability to get ahead in our country these days?\newline
 \newline
A. Helps a lot\newline
B. Hurts a lot\newline
C. Neither helps nor hurts\newline
D. Refused\newline
E. Helps a little\newline
F. Hurts a little\newline
Answer: & This question seems to be asking about the potential impact of being an evangelical Christian on a person's success or social mobility in the United States. The options provided suggest that the impact could be positive, negative, or neutral. However, it's important to note that the impact of evangelical Christianity on a person's ability to get ahead can vary widely depending on a number of factors, including the individual's personal background, the 
specific community they live in, and the attitudes and biases of the people around them.
  & nan\\

\bottomrule
\end{tabular}
}
\caption{Different cases of model outputs. }
\label{tab:examples}
\end{table*}
\end{document}